\documentclass[5p]{elsarticle} %双栏
\usepackage{amssymb}
\usepackage{float}
\usepackage{xcolor}
\usepackage{array}
\usepackage{booktabs}
\usepackage{bbding}
\usepackage{colortbl}
\usepackage{lineno}
\usepackage{multirow}
\usepackage{amsmath,amssymb,amsfonts}
\usepackage{algorithmic}
\usepackage{graphicx}
\usepackage{textcomp}
\usepackage{color}
\journal{Neural Network}

\usepackage{ecrc}
\volume{00}
\firstpage{1}
\runauth{}
\jid{procs}
\jnltitlelogo{Neural Network}
\CopyrightLine{2024}{Published by Elsevier Ltd.}

\newcommand{\colorbibs}[2][blue]%
{%
	\DeclareBibliographyCategory{ColoredBiblist#1}%
	\addtocategory{ColoredBiblist#1}{#2}%
	\AtEveryBibitem{\ifcategory{ColoredBiblist#1}{\color{#1}\bfseries}{}}
}

\begin{document}
%\linenumbers

\begin{frontmatter}



\title{Contrastive Graph Representation Learning with Adversarial Cross-view
	Reconstruction and Information Bottleneck}


\author[address2]{Yuntao Shou$^{\mathrm{a }, }$}
\ead{shouyuntao@stu.xjtu.edu.cn}

\author[address3]{Haozhi Lan}
\ead{haozhilan1@gmail.com}

\author[address2]{Xiangyong Cao$^{\mathrm{a }, }$\corref{cor1}}
\ead{caoxiangyong@mail.xjtu.edu.cn}
\cortext[cor1]{Corresponding author}%corresponding author

\address[address1]{School of Computer Science and Technology, Xi’an Jiaotong University, Xi’an, China}

\address[address2]{Ministry of Education Key Laboratory for Intelligent Networks and Network Security, Xi’an Jiaotong University, Xi’an, China}

\address[address2]{School of Artificial Intelligence, Xi'an Jiaotong University}

\begin{abstract}
Graph Neural Networks (GNNs) have received extensive research attention due to their powerful information aggregation capabilities. Despite the success of GNNs, most of them suffer from the popularity bias issue in a graph caused by a small number of popular categories. Additionally, real graph datasets always contain incorrect node labels, which hinders GNNs from learning effective node representations. Graph contrastive learning (GCL) has been shown to be effective in solving the above problems for node classification tasks. Most existing GCL methods are implemented by randomly removing edges and nodes to create multiple contrasting views, and then maximizing the mutual information (MI) between these contrasting views to improve the node feature representation. 
However, maximizing the mutual information between multiple contrasting views may lead the model to learn some redundant information irrelevant to the node classification task. To tackle this issue, we propose an effective Contrastive Graph Representation Learning with Adversarial Cross-view
Reconstruction and Information Bottleneck (CGRL) for node classification, which can adaptively learn to mask the nodes and edges in the graph to obtain the optimal graph structure representation. Furthermore, we innovatively introduce the information bottleneck theory into GCLs to remove redundant information in multiple contrasting views while retaining as much information as possible about node classification. Moreover, we add noise perturbations to the original views and reconstruct the augmented views by constructing adversarial views to improve the robustness of node feature representation. Extensive experiments on real-world public datasets demonstrate that our method significantly outperforms existing state-of-the-art algorithms.
\end{abstract}

\begin{keyword}
Graph Neural Networks, Contrastive Learning, Mutual Information, Information Bottleneck, Adversarial Learning
\end{keyword}

\end{frontmatter}

\section{Introdution}
Graph Neural Networks (GNNs) have attracted extensive attention from researchers due to the increase in large amounts of real-world graph-structured data \cite{xu2020radial, zhou2018graph, shou2022conversational, shou2023comprehensive, meng2023deep, meng2024multi, shou2023adversarial, ai2024gcn}. Meanwhile, GNNs are widely used in intelligent recommender systems, and social media fields because they provide a practical way to aggregate high-order neighbor information \cite{he2020lightgcn, wang2019neural, ai2023two, shou2023masked, shou2022object}.

\begin{figure}
	\centering
	\includegraphics[width=1\linewidth]{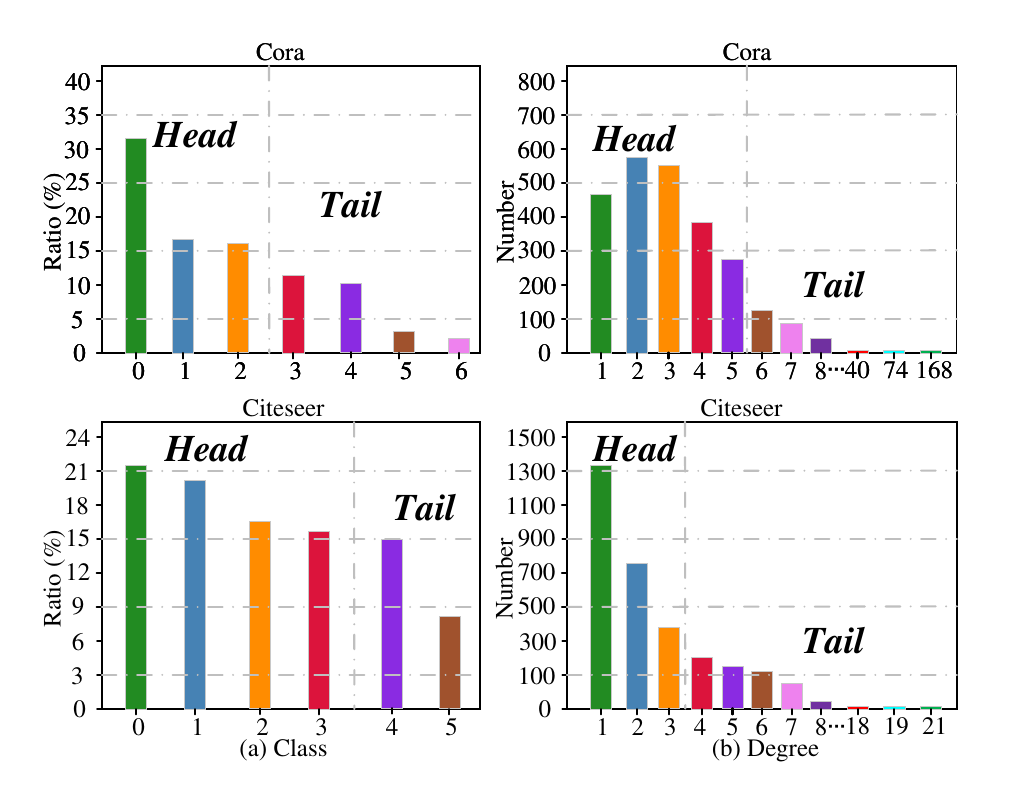}
	\caption{(a) The proportion of papers in different categories on the cora and citeseer datasets. (b) The degree distribution of different nodes on the cora and citeseer datasets. By analyzing the distribution of categories and degrees, we argue that the constructed graph structure has a long-tail problem.}
	\label{fig:longtail}
\end{figure}

Although GNNs have achieved reliable performance on node classification tasks, we argue that most of the node classification models based on GNNs suffer from the following two problems. i) \textbf{Popularity Bias.} As shown in Fig. \ref{fig:longtail}, different categories of papers have different numbers, and the number of citations of papers also varies, and this unbalanced learning can lead to the popularity bias problem in GNNs learning. In most node classification tasks, the categories and degrees of nodes follow a long-tail distribution, which means that popular categories have many papers, while most papers have few citations. In other words, most of the nodes have less interaction, which hinders the information update of the nodes. The above-mentioned popularity bias problem cause GNNs to tend to learn node representations that are popular and have many interactions, which hinders the representation learning of GNNs. ii) \textbf{Noise Interference.} There may be miscitations in the citation process of papers (i.e., there is a citation relationship between two unrelated papers in different fields), which leads to noise in the information contained in the data. Studies have shown that the feature extraction ability of GCN is closely related to the quality of the input graph, which means that the input image with noise may cause the model to learn poor solutions. In response to the above problems, existing graph contrastive learning (GCL) methods \cite{velivckovicdeep, shou2023low, ai2023gcn, shou2024adversarial}, \cite{zhu2021graph}, \cite{wu2021self}, \cite{yu2021self}, \cite{xia2021self} propose an effective solution mechanism to alleviate popular bias and improve the robustness of the GCN model.

Nonetheless, the above-mentioned methods suffer from two limitations. 1) Most GCL methods perform data augmentation to optimize the graph structure by randomly masking nodes or perturbing edges. However, the strategy of randomly masking nodes and edge perturbations is too random, which may cause serious damage to the semantic information of the graph structure. For example, in the functional prediction of molecular structural properties, if edges are randomly perturbed, the structural properties of the molecule will change greatly. In addition, the interpretability that can alleviate popular paranoia and improve model robustness through the above methods is relatively poor. 2) The purpose of existing GCL methods \cite{bachman2019learning, shou2023graph, shou2023czl}, \cite{peng2020graph}, \cite{tschannen2019mutual} to generate multiple views through data augmentation is to maximize the mutual information between views, which may cause the model to capture task-irrelevant feature information. Inspired by the information bottleneck theory \cite{tishby1999information}, we believe that a good GCL method should reduce as much redundant information as possible while retaining as much task-related information as possible.

To tackle the aforementioned issues, we propose a novel method called Contrastive Graph Representation Learning with Adversarial Cross-view
Reconstruction and Information Bottleneck (CGRL) for node classification. CGRL consists of two key components: i.e., adaptive automatic generation of graph-augmented views and graph contrastive learning via information bottlenecks.

First, this paper designs an automatic graph augmentation that adaptively learns node masks and edge perturbations to optimize the original graph into relevant views. In addition, SCGCL employs a joint training strategy to train an adaptive learnable view generator and node classifier in an end-to-end manner, thereby generating augmented views with structural heterogeneity but semantic similarity. As a result, the generated augmented view can undersample the popular nodes in the original graph while retaining the majority of isolated nodes to alleviate the model's popularity bias problem. Intuitively, random masking nodes or perturbed edges do not consider the distribution probabilities and neighborhood information of different types of nodes in the original graph, but dropout them randomly. However, GCN based on message passing is difficult to reconstruct the information of isolated nodes and it is easier to optimize the semantic information of popular nodes. Therefore, the model may achieve better classification results on popular nodes and poor classification results on isolated nodes. The method CGRL proposed in this paper takes the augmented views of debias information and inputs them into GCN for node classification, which improves the model's ability to resist popular bias.

Second, we integrate multiple views that are semantically similar and contain complementary information into a shared feature space for compact representation, which can improve the robustness of CGRL. The intuition behind is that when different views contain complementary semantic information, the model can obtain more prior knowledge to improve the performance of node classification tasks \cite{wan2021multi, shou2024efficient, shou2024revisiting, meng2024masked}. However, we argue that maximizing mutual information between different views forces the model to learn redundant information that is irrelevant to downstream tasks. Inspired by the information bottleneck (IB) theory \cite{yu2022improving}, it obtains optimal solutions by maximizing label information relevant to downstream tasks and minimizing mutual information between different views. Based on the IB strategy, an automatic graph augmenters learns to generate augmented views that remove noise information and contain semantically similar and complementary views. In addition, when calculating the contrastive loss, we not only use the node feature representations of the two augmented views, but also introduce the node feature representations of the original view perturbed in an adversarial manner as a third view. This additional adversarial view introduces perturbations that force the model to not only accurately distinguish the semantic features of the augmented views, but also to maintain an understanding of the semantic integrity of the original graph in the face of perturbations. Through multi-view adversarial reconstruction, we further improve the robustness of the feature representations.

Compared with the previous methods, the contributions of this paper method are summarized as:

Firstly, we propose a Contrastive Graph Representation Learning with Adversarial Cross-view Reconstruction and Information Bottleneck (CGRL) for node classification. The CGRL method can alleviate the popularity bias and interaction noises problem of the existing GNNs in aggregating neighbor node information.

Secondly, The CGRL approach provides a learnable approach to adaptively mask nodes and edges for multiple graph contrastive views in an end-to-end manner. In addition, redundant information irrelevant to node classification is discarded by innovatively introducing information bottleneck theory into multi-view graph contrastive learning.

Thirdly, we introduce a cross-view adversarial reconstruction strategy to further improve the robustness of node feature representation. 

Finally, extensive experiments also show that the proposed CGRL method outperforms the state-of-the-art methods on seven real-world publicly available datasets. 

\section{Related Work}

\textbf{Graph Representation Learning} \quad
Early work \cite{ou2016asymmetric, ji2015knowledge, nie2017unsupervised} have made great progress in representation learning tasks (e.g., node classification, entity alignment, and link prediction, etc). Specifically, GNNs on graph representation learning usually follow the information aggregation mechanism to update the feature representation of nodes, i.e., stimulating connected neighbor nodes to have similar semantic information. Inspired by GNNs, several works (e.g., AROPE \cite{zhang2018arbitrary}, DeepWalk \cite{perozzi2014deepwalk}, and GraphSAGE \cite{hamilton2017inductive}, etc) on graph representation learning usually follows the label propagation mechanism to update the feature representation of nodes, i.e., stimulating connected neighbor nodes to have similar labels. In recent years, GNNs (e.g., GraphSMOTE \cite{zhao2021graphsmote}, Nodeformer \cite{wu2022nodeformer}, DisGNN \cite{zhao2022exploring}, and GraphFL \cite{wang2022graphfl}, etc) have applied GNN algorithms to learn discriminative latent representations on node classification tasks \cite{shou2023graphunet, meng2024revisiting}.

\textbf{Contrastive Learning} \quad
Contrastive learning (CL) \cite{xu2019ternary, tian2020contrastive, hjelmlearning}, which is originally widely used in computer vision to obtain better image feature representations, has received extensive research attention in graph learning. Specifically, graph contrastive learning (GCL) learns discriminative node representations by maximizing mutual information (MI) between multiple graph views. For instance, DGI \cite{velivckovicdeep} learns more discriminative node representations by contrasting node embeddings of local and global graph views. GIC \cite{mavromatis2021graph} maximizes the MI between node clusters with high similarity to make full use of coarse-grained and fine-grained information between nodes. CMC \cite{tian2020contrastive} maximizes MI by contrasting feature representations from different views. GMI \cite{peng2020graph} estimates and maximizes the MI between the input graph and the feature representation from two aspects of node features and network topology. Some recent self-supervised graph learning (SGLs) methods (e.g., MGAE \cite{tan2022mgae}, GraphMAE \cite{hou2022graphmae}, and GAE \cite{li2022maskgae}) generate multiple graph views of nodes and edges and force the consistency between different graph views. On the one hand, all these methods generate multiple contrasting graph views by randomly masking nodes or edges, which may cause some important structural and semantic information to be lost. On the other hand, these methods maximize mutual information between different graph views, which may force the model to learn some task-independent semantic information. To sum up, existing self-supervised graph learning for node classification suffer from insufficient information utilization, i.e., information associated with the label.

\begin{figure*}
	\centering
	\includegraphics[width=1\linewidth]{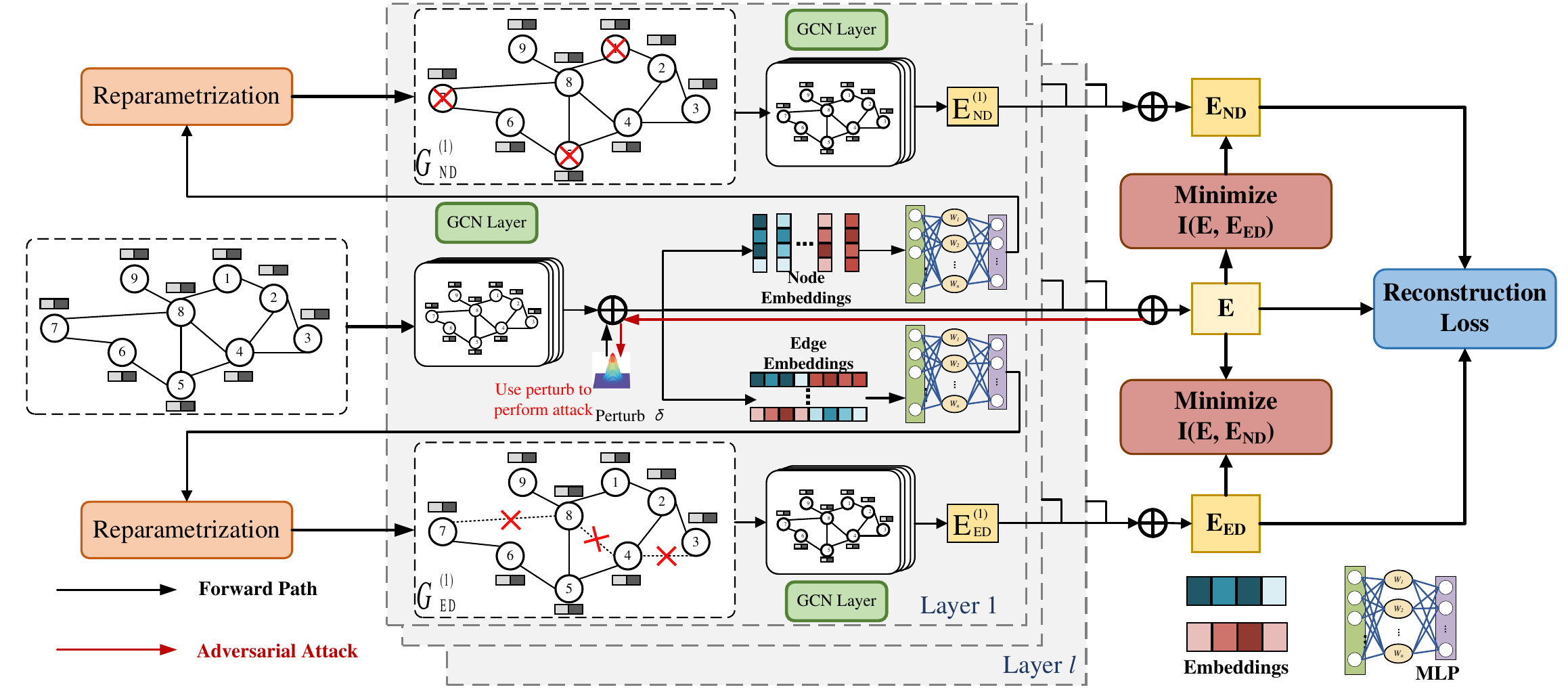}
	\caption{The overview of the proposed CGRL method. Specifically, we combine node dropout and edge dropout views to obtain a more comprehensive representation, where node dropout view can alleviate the problem of popular bias, and edge dropout can alleviate the problem of misconnection between nodes.}
	\label{fig:architecture}
\end{figure*}

\textbf{Information-Bottleneck Representation Learning} \quad
The information bottleneck (IB) theory \cite{tishby1999information} argues that if the feature representation learned by the model from the input data discards information that is not useful for the given task while retaining as much as possible the information relevant to the given task, it will increase the generalization of downstream tasks. Formally, IB needs to construct multiple views for feature representation learning. Motivated by the effectiveness of the IB, several works have considered transferring information bottleneck theory to graph representation learning tasks. For instance, MIB \cite{federici2020learning} designs an unsupervised multi-view method and uses the information bottleneck theory to minimize the representation of redundant information.  DeepIB \cite{wang2019deep} reduces redundant information irrelevant to a given task by minimizing mutual information between multiple views and input data. CMIB \cite{wan2021multi} combines information bottleneck theory to capture the complementarity of intrinsic information between different views and balance the consistency of multi-view latent representations. Nevertheless, almost all the above-mentioned methods aim to obtain a discriminative graph view to replace the original input graph, which may cause some semantic information and topological information of the original graph to be lost. In conclusion, existing information bottleneck methods for node classification suffer from insufficient information utilization, i.e., input graph information and multi-view information.

\section{Preliminaries}

\textbf{Notations.}
Suppose $\mathcal{G}=\{\mathcal{V}, \mathcal{E},\mathcal{R},\mathcal{W}\}$ represents a graph where $\mathcal{V}=\{v_1, v_2, \ldots, v_M\}$ is the nodes set, $M$ is the number of nodes, $\mathcal{E} \subseteq \mathcal{V} \times \mathcal{V}$ represent the edges set, $r_{ij}(r_{ij} \in \mathcal{E})$ represents the connection relationship between node $i$ and node $j$, and $\omega_{ij} (\omega_{ij} \in \mathcal{W},0\leq \omega_{ij} \leq 1)$ the weight of the edge $\mathcal{E}_{ij}$. The feature matrix and degree matrix of nodes are expressed as $X= \{x_i\}_{i=1}^{M}$ amd $A=\{a_{ij}\} \in \{0, 1\}^{M \times M}$, respectively, where $x_i$ represent the features of the node $v_i$, and if $(v_i, v_j) \in \mathcal{E}$ then $a_{ij}=1$ otherwise $a_{ij}=0$. $\mathcal{G} = G(x)$ is regarded as the process of graph processing.

\textbf{GCN Processing.} For input features $X \in \mathbb{R}^{M \times D}$, a graph $\mathcal{G} = F=G(X)$ is constructed based on the features $X$. A GCN layer is utilized to update features representations between nodes by aggregating information from their neighbor nodes. Specially, GCN operates as follows:
\begin{gather}
	\begin{aligned}
		\mathcal{G}^{\prime} & =F(\mathcal{G}, \mathcal{W}) \\
		& = Update\left({Aggregate}\left(\mathcal{G}, W_{agg}\right), W_{ {update}}\right) \\
		& = W_{update}\left(\sum_{r \in \mathcal{R}}\sum_{j \in N_i^r} \frac{1}{|\mathcal{N}_i^r |} \left(\omega_{ij} W_{agg} \mathcal{G}+\omega_{ii} W_{agg} \mathcal{G}\right)\right)
	\end{aligned}
\end{gather}
where $\mathcal{N}_i^r$ is the set of neighbor nodes of node $i$ under the edge relationship $r \in \mathcal{R}$, and $W_{agg}$ and $W_{update}$ represent the learnable weights of the nodes in aggregating surrounding neighbor nodes information and updating the aggregated features, respectively.

In addition, we also introduce a multi-head attention mechanism in the GNN layer to capture node feature information and topology information in a more fine-grained manner. The feature vectors $x_i^{''}$ after being aggregated and updated is divided into $N$ heads, i.e., $h^1, h^2, \ldots, h^N$, and each head is assigned a learnable parameter. Therefore, the feature vectors $x$ is finally updated as follows:
\begin{gather}
	\mathbf{x}_i^{\prime}=\left[h^1 W_{update}^1, h^2 W_{update}^2, \ldots, h^N W_{update}^N\right],
\end{gather}

\textbf{Information Bottleneck (IB).} IB is an information theory-based strategy that describes the information in the data that is relevant to downstream tasks. IB argues that if the obtained feature representation excludes semantic information in the original input that is irrelevant to a given downstream task, it improves the robustness of the model. Specifically, for a given input data $x$, with associated label information is $y$, using the IB strategy for model optimization can obtain a compact feature representation $z$. The optimization goals of IB are as follows:
\begin{gather}
	\label{eq:3}
	\max _{\mathbf{Z}} I(\mathbf{Y}, \mathbf{Z}, \mathbf{\theta})-\beta I(\mathbf{X}, \mathbf{Z}, \mathbf{\theta}),
\end{gather}
where $\beta$ is an adjustment factor, $\theta$ is a learnable parameter.

Combining $I(Y,Z)$ and mutual information theory in Eq. \ref{eq:3}, we can get:
\begin{gather}
	\begin{aligned}
		I(\mathbf{Y}, \mathbf{Z}) & =\int d \mathbf{y} d \mathbf{z} p(\mathbf{y}, \mathbf{z}) \log \frac{p(\mathbf{y}, \mathbf{z})}{p(\mathbf{y}) p(\mathbf{z})} \\
		& =\int d \mathbf{y} d \mathbf{z} p(\mathbf{y}, \mathbf{z}) \log \frac{p(\mathbf{y} \mid \mathbf{z})}{p(\mathbf{y})}.
	\end{aligned}
\end{gather}

However, directly calculating $p(y|z)$ is quite difficult. Inspired by \cite{alemi2016deep}, we use $q(y|z)$ as the variational approximation of $p(y|z)$. 

Since the $KL$ divergence is always greater than or equal to 0, we can get:
\begin{gather}
	\begin{aligned}
		& \mathbf{K L}[p(\mathbf{y} \mid \mathbf{z}), q(\mathbf{y} \mid \mathbf{z})] \geq 0 \Rightarrow \int d \mathbf{y} p(\mathbf{y} \mid \mathbf{z}) \log \frac{p(\mathbf{y} \mid \mathbf{z})}{q(\mathbf{y} \mid \mathbf{z})} \geq 0 \\
		\Rightarrow & \int d \mathbf{y} p(\mathbf{y} \mid \mathbf{z}) \log p(\mathbf{y} \mid \mathbf{z}) \geq \int d \mathbf{y} p(\mathbf{y} \mid \mathbf{z}) \log q(\mathbf{y} \mid \mathbf{z}).
	\end{aligned}
\end{gather}

Therefore we can know:
\begin{gather}
	\begin{aligned}
		I(\mathbf{Y}, \mathbf{Z}) & \geq \int d \mathbf{y} d \mathbf{z} p(\mathbf{y}, \mathbf{z}) \log \frac{q(\mathbf{y} \mid \mathbf{z})}{p(\mathbf{y})} \\
		& =\int d \mathbf{y} d \mathbf{z} p(\mathbf{y}, \mathbf{z}) \log q(\mathbf{y} \mid \mathbf{z})+H(\mathbf{y}) \\
		& \geq \int d \mathbf{y} d \mathbf{z} p(\mathbf{y}, \mathbf{z}) \log q(\mathbf{y} \mid \mathbf{z}) \\
		& =\int d \mathbf{y} p(\mathbf{y}) \int d \mathbf{z} p(\mathbf{z} \mid \mathbf{y}) \log q(\mathbf{y} \mid \mathbf{z}) .
	\end{aligned}
\end{gather}

For $I(X,Z)$ in Eq. \ref{eq:3}, we can get:
\begin{gather}
	\begin{aligned}
		I\left(\mathbf{Z}, \mathbf{X}\right)&= 	\int d \mathbf{z} d \mathbf{x} p\left(\mathbf{x}, \mathbf{z}\right) \log \frac{p\left(\mathbf{z} , \mathbf{x}\right)}{p(\mathbf{z}), p(\mathbf{x})}  \\
		&= \int d \mathbf{z} d \mathbf{x} p\left(\mathbf{x}, \mathbf{z}\right) \log \frac{p\left(\mathbf{z} \mid \mathbf{x}\right)}{p(\mathbf{z})}.
	\end{aligned}
\end{gather}

Similarly, it is quite difficult to calculate $p(z)$. We use $r(z)$ as the variational approximation to estimate $p(z)$. Since $\mathbf{K L}[p(\mathbf{z}), r(\mathbf{z})] \geq 0 \Rightarrow \int d \mathbf{z} p\mathbf{(z)} \log p(\mathbf{z}) \geq \int d \mathbf{z} p\mathbf{(z)} \log r(\mathbf{z})$, we can get an upper bound:
\begin{gather}
	\begin{array}{r}
		I\left(\mathbf{Z}, \mathbf{X}\right) \leq \int d \mathbf{x} d \mathbf{z} p\left(\mathbf{x}\right) p\left(\mathbf{z} \mid \mathbf{x}\right) \log \frac{p\left(\mathbf{z} \mid \mathbf{x}\right)}{r(\mathbf{z})} \\
		=\int d \mathbf{x} p\left(\mathbf{x}\right) \int d \mathbf{z} p\left(\mathbf{z} \mid \mathbf{x}\right) \log \frac{p\left(\mathbf{z} \mid \mathbf{x}\right)}{r(\mathbf{z})} .
	\end{array}
\end{gather}

Combining the above analysis and inequalities, we can obtain the lower bound of the information bottleneck theory:
\begin{gather}
	\begin{aligned}
		& I(\mathbf{Y}, \mathbf{Z})-\sum_{v=1}^V \beta I\left(\mathbf{Z}, \mathbf{X}\right) \\
		& \geq \int d \mathbf{y} p(\mathbf{y}) \int d \mathbf{z} p(\mathbf{z} \mid \mathbf{y}) \log q(\mathbf{y} \mid \mathbf{y}) \\
		& - \beta \int d \mathbf{x} p\left(\mathbf{x}\right) \int d \mathbf{z} p\left(\mathbf{z} \mid \mathbf{x}\right) \log \frac{p\left(\mathbf{z} \mid \mathbf{x}\right)}{r(\mathbf{z})}.
	\end{aligned}
\end{gather}

\section{Methodology}
In this section, we introduce our proposed Contrastive Graph Representation Learning with Adversarial Cross-view Reconstruction and Information Bottleneck (CGRL) method in detail.

\subsection{Automatically Generated Multi-view Augmentation}

\textbf{Adversarial View.} By adding perturbed adversarial examples to the original image views, the robustness of the model is significantly enhanced. This improvement can be attributed to a possible explanation that there is still non-predictive redundant information in the information shared between the two augmented views. When adversarial examples are introduced, this redundant information is weakened or eliminated, forcing the model to focus on more important and discriminative features. We define adversarial view as follows:
\begin{gather}
		\mathcal{G}_{adv}^{(l)}=\mathcal{G} + \delta^* 
\end{gather}
\begin{gather}
	\begin{aligned}
	\delta^*&=\underset{\left\|\delta\right\|_\infty\leqslant\epsilon}{\mathrm{argmax}}\mathcal{L}_{\mathrm{adv}}\left(\mathcal{G}_{E D},\mathcal{G}_{N D},\mathcal{G}+\delta\right) \\
	&=\underset{\left\|\delta\right\|_\infty\leqslant\epsilon}{\mathrm{argmax}}\max_{\delta^{*}}\left[\mathcal{L}_{\mathrm{CL}}\left(\mathcal{G}_{N D},G+\delta^{*}\right) \right.\\
	&\left.+\mathcal{L}_{\mathrm{CL}}\left(\mathcal{G}_{E D},G+\delta^{*}\right)\right]
	\end{aligned}
\end{gather}
where $\mathcal{G}_{adv}^{(l)} = \{\mathcal{V}', \mathcal{E}',\mathcal{R}',\mathcal{W}'\}$, $\mathcal{V}=\{v_1, v_2, \ldots, v_M\}$, $\delta$ is the randomly initialized Gaussian noise, $\mathcal{L}_{CL}$ is the infoNCE loss, $\epsilon$ is the radius. Inspired by recent work \cite{yang2021graph}, we add a perturbation $\delta$ to the output of the first hidden layer. It has been empirically shown that it can more effectively perturb the intermediate representation of the model than adding perturbations to the initial node features, allowing the model to learn and predict in more complex environments.

\textbf{Node-Masking View.}
As shown in Fig. 1, both the category of nodes and the degree of nodes in citation data show data imbalance, which hinders GCN from learning the feature representation of minority class nodes. Therefore, we perform automatic learnable node masking before each information aggregation and feature update of GCN to alleviate the shielding effect of influential nodes on minority class nodes. The node-masking view we created is formulated as follows:
\begin{gather}
	\mathcal{G}_{N D}^{(l)}=\left\{\left\{v_i' \odot \eta_i^{(l)} \mid v_i' \in \mathcal{V}'\right\}, \mathcal{E}', \mathcal{R}', \mathcal{W}'\right\}
\end{gather}
where $\eta_i^{(l)} \in \{0,1\}$ is sampled from a parameterized Bernoulli distribution $Bern(\omega_i^l)$, and $\eta_i^{(l)} = 0$ represents masking node $v_i$, $\eta_i^{(l)} = 1$ represents keeping node $v_i$.

Randomly removing some nodes and their connections in the graph may result in a large loss of minority class node information, thereby affecting the information aggregation of minority class nodes and leading to unsatisfactory classification results. Therefore, instead of directly removing the selected nodes from the graph, we replace the selected nodes by sampling the representation of the local subgraph using a random walk strategy to obtain a local representation of the node.

\textbf{Edge Perturbation View.}
The goal of perturbing edges is to generate an optimized graph structure that filters noisy edges and alleviates the problem of popularity bias. The edge perturbation view is formulated as follows:
\begin{gather}
	\mathcal{G}_{E D}^{(l)}=\left\{\mathcal{V}',\left\{e_{i j}' \odot \eta_{i j}^{(l)} \mid e_{i j}' \in \mathcal{E}', \mathcal{R}', \mathcal{W}'\right\}\right\}
\end{gather}
where $\eta_
{ij}^{(l)} \in \{0,1\}$ is also sampled from a parameterized Bernoulli distribution $Bern(\omega_{ij}^l)$, and $\eta_{ij}^{(l)} = 0$ represents perturbating edges $e_{ij}$, $\eta_i^{(l)} = 1$ represents keeping edge $e_{ij}$.

To enable the model to automatically learn whether to mask nodes and perturb edges, we formally define the learnable parameter $\omega^l_i$ and $\omega^l_{ij}$ as follows:
\begin{gather}
	\omega_i^{(l)}=\operatorname{Linear}\left(\mathbf{e}_i'^{(l)}\right) ; \quad \omega_{i j}^{(l)}=\operatorname{Linear}\left(\left[\mathbf{e}_i'^{(l)} ; \mathbf{e}_j'^{(l)}\right]\right)
\end{gather}

To ensure that the model can automatically optimize and generate augmented multi-views in an end-to-end learning method, we use reparameterization technology \cite{jang2016categorical} to convert the discretized $\eta$ into a continuous function. The formula is defined as follows:
\begin{gather}
	\eta_i=\frac{\exp \left(\left(\log \left(\pi_i\right)+g_i\right) / \tau\right)}{\sum_{j=1}^m \exp \left(\left(\log \left(\pi_j\right)+g_j\right) / \tau\right)}, \quad \text { for } i=1, \ldots, m
\end{gather}
where $g_i=-log\left(-log\left(\epsilon_i\right)\right), \epsilon_i \sim Uniform(0, 1)$, $\tau \in \mathbb{R}^+$ means annealing temperature, $\tau$ represents the class probability, and $m$ represents number of categories.

After obtaining the masked node and edge perturbed views, we input them into GCN for feature representation to obtain optimized multi-views. The formula is defined as follows:
\begin{gather}
	\begin{aligned}
		&\mathbf{E}_{N D}^{(l)}=GraphConv\left(\mathbf{E}_{N D}^{(l-1)}, \mathcal{G}_{N D}^{(l)}\right) \\ 
		&\mathbf{E}_{E D}^{(l)}=GraphConv\left(\mathbf{E}_{E D}^{(l-1)}, \mathcal{G}_{E D}^{(l)}\right)
	\end{aligned}
\end{gather}
where $GraphConv$ represents the graph convolution operation, and we choose GAT as our graph encoder. $\mathbf{E}_{N D}$ and $\mathbf{E}_{E D}$ represent the node feature representations of node-masking view and edge perturbation view respectively, $\mathbf{G}_{N D}$ and $\mathbf{G}_{E D}$ represent node-masking view and edge perturbation view respectively.

\subsection{Contrastive Learning via IB}
Although CGRL combines an automatic learnable view augmentation and a node classification process for model optimization, we argue that relying solely on the classification objective does not well guide the node masking and edge perturbation process to create optimal multi-views. Therefore, we follow the information bottleneck strategy \cite{tishby1999information} to retain sufficient semantic information relevant to downstream tasks in the augmented node mask and edge perturbation views. Specifically, unlike traditional CL strategies, we encourage maintaining topological heterogeneity between the augmented view and the original graph while maximizing the information relevant to the node classification task. Based on the above strategy, we can obtain topologically heterogeneous but semantically similar enhanced multi-view representations and effectively remove noise information in the graph. Therefore, the objective in Eq. \ref{eq:3} is summarized as:
\begin{gather}
	\begin{aligned}
		& \min_{(E, \tilde{E})} \mathcal{L}_{CLS} + I(\mathbf{E}, \tilde{\mathbf{E}}) \\
		& = -\sum_{i=1}^{n}y_ilog(\hat{y}) + I(\mathbf{E}, \tilde{\mathbf{E}})
	\end{aligned}
\end{gather}
where $L_{CLS}$ represents the cross-entropy loss, $I(\mathbf{E}, \tilde{\mathbf{E}})$ represents the mutual information between the original view and the augmented view.

Following \cite{oord2018representation, sun2019infograph}, minimizing the lower bound of mutual information (i.e., Eq. 9) is equivalent to maximizing the InfoNCE loss \cite{gutmann2010noise}. Therefore, we adopt negative InfoNCE to optimize the feature representation between the augmented view and the original graph. Specifically, we treat the same node representations in the original graph and the automatically generated view as positive pairs (i.e., $\{(e_i', \tilde{e}_i')|v_i' \in \mathcal{V}'\}$), and different node representations as negative pairs (i.e., $\{(e_i', \tilde{e}_j')|v_i', v_j' \in \mathcal{V}'\}, i \neq j$).
\begin{gather}
	\begin{aligned}
		I\left(\mathbf{E}, \tilde{\mathbf{E}}\right) & = \int d \mathbf{y} p(\mathbf{y}) \int d \mathbf{z} p(\mathbf{z} \mid \mathbf{y}) \log q(\mathbf{y} \mid \mathbf{y}) \\
		& - \beta \int d \mathbf{x} p\left(\mathbf{x}\right) \int d \mathbf{z} p\left(\mathbf{z} \mid \mathbf{x}\right) \log \frac{p\left(\mathbf{z} \mid \mathbf{x}\right)}{r(\mathbf{z})} \\
		& = \sum_{v_i' \in \mathcal{V}'} \log \frac{\exp \left(sim\left(\mathbf{e}_i', \tilde{\mathbf{e}}_i'\right) / \tau\right)}{\sum_{v_j \in \mathcal{V}} \exp \left(sim\left(\mathbf{e}_i', \tilde{\mathbf{e}}_j'\right) / \tau\right)}.
	\end{aligned}
\end{gather}
where $s(\cdot)$ is used to measure the similarity between positive and negative sample pairs.

\subsection{Adversarial Cross-view Reconstruction}
To further achieve feature disentanglement, we propose a cross-view reconstruction mechanism. Specifically, we hope that the representation pairs within and across enhanced views can recover the original data. More specifically, we define ($\mathbf{E}_{ND}$, $\mathbf{E}_{ED}$), ($\mathbf{E}$, $\mathbf{E}_{ED}$), ($\mathbf{E}$, $\mathbf{E}_{ED}$) as a cross-view representation pair, and repeat the reconstruction process on it to predict the original view, aiming to ensure that $\mathbf{E}_{ND}$, $\mathbf{E}_{ED}$, $\mathbf{E}$ are optimized to approximately disentangle each other. Intuitively, the reconstruction process is able to separate the information of the shared feature set from the information in the unique feature set between the two enhanced views. We formally define the reconstruction process as:
\begin{gather}
	\begin{aligned}
	\mathcal{L}_{\mathrm{recon}}=&\frac{1}{2N}\left[\left\|\mathbf{E}-\mathbf{E}_{ND}\right\|_{2}^{2}+\left\|\mathbf{E}-\mathbf{E}_{ED}\right\|_{2}^{2}\right]
	\end{aligned}
\end{gather}
wherer $N$ represents the number of nodes.

\subsection{Model Training}
We train the model with the goal of optimizing the node mask view, edge perturbation view, and node classification:
\begin{gather}
	\begin{aligned}
		\mathcal{L}&=\mathcal{L}_{CLS}+\alpha\mathcal{L}_{\text {CLS }}^{N D}
		+\left(1-\alpha\mathcal{L}_{\text {CLS }}^{E D}\right)\\
		&+\beta\left(I\left(\mathbf{E}, \mathbf{E}_{N D}\right)+I\left(\mathbf{E}, \mathbf{E}_{E D}\right)\right)
		+\mathcal{L}_{\mathrm{recon}}+\lambda\|\Theta\|_2^2
	\end{aligned}
\end{gather}
where $\lambda$ and $\beta$ represent the learning weights of L2 regularization and information bottleneck contrast learning respectively.

\section{Theoretic Analysis} 
\textbf{Theorem 1.} Specifically, we regard the augmented view as $\tilde{\mathcal{G}}$, the noisy view as $\mathcal{G}^{\prime}$, and the node label information as $Y_{\operatorname{CLS}}$. We assume that $\hat{\mathcal{G}}$ is not related to the classification information $Y_{\operatorname{CLS}}$. Therefore, the upper bound of $I\left(\hat{\mathcal{G}} ; \tilde{\mathcal{G}}\right)$ is defined as follows:
\begin{gather}
	\label{eq.18}
	I\left(\hat{\mathcal{G}} , \tilde{\mathcal{G}}\right) \leq I(\mathcal{G} , \tilde{\mathcal{G}})-I\left(Y_{\operatorname{CLS}} , \tilde{\mathcal{G}}\right)
\end{gather}

\textbf{\textit{Proof.}} We assume that the clean graph $\mathcal{G}$ is defined by $\mathcal{G}^\prime$ and label information $Y$, and we can get $(\mathcal{G}^\prime, Y_{CLS}) \rightarrow \mathcal{G} \rightarrow \tilde{\mathcal{G}}$ according to the Markov chain \cite{achille2018emergence}. Therefore, we can get:
\begin{gather}
	\label{eq.19}
	\begin{aligned}
		I(\mathcal{G} , \tilde{\mathcal{G}}) &\geq I\left(\left(Y_{\operatorname{CLS}}, \hat{\mathcal{G}}\right) , \tilde{\mathcal{G}}\right) \\ & =I\left(\hat{\mathcal{G}} , \tilde{\mathcal{G}}\right)+I\left(Y_{\operatorname{CLS}} , \tilde{\mathcal{G}} \mid \hat{\mathcal{G}}\right) \\
		& =I\left(\hat{\mathcal{G}}, \tilde{\mathcal{G}}\right)+H\left(Y_{\operatorname{CLS}} \mid \hat{\mathcal{G}}\right)-H\left(Y_{\operatorname{CLS}} \mid \hat{\mathcal{G}}, \tilde{\mathcal{G}}\right)
	\end{aligned}
\end{gather}

Because there is no correlation between $\tilde{\mathcal{G}}$ and $Y_{\operatorname{CLS}}$, $H(Y_{CLS} \mid \tilde{\mathcal{G}}) = H(Y_{CLS})$. Furthermore, $H\left(Y_{\operatorname{CLS}} \mid \hat{\mathcal{G}}, \tilde{\mathcal{G}}\right) \leq H\left(Y_{\operatorname{CLS}}\right)$. Therefore, we simplify the Eq. \ref{eq.19} as follows:
\begin{gather}
	\begin{aligned}
		I(\mathcal{G} , \tilde{\mathcal{G}}) &\geq  I\left(\hat{\mathcal{G}} , \tilde{\mathcal{G}}\right)+H\left(Y_{\operatorname{CLS}}\right)-H\left(Y_{\operatorname{CLS}} \mid \tilde{\mathcal{G}}\right) \\&=   I\left(\hat{\mathcal{G}}, \tilde{\mathcal{G}}\right)+I\left(Y_{\operatorname{CLS}} , \tilde{\mathcal{G}}\right)
	\end{aligned}
\end{gather}
Therefore, we prove that Eq. \ref{eq.18} holds. In summary, we provide a theoretical basis to ensure that graph contrastive learning via information bottlenecks can achieve noise invariance by reducing redundant and interfering information in augmented views.

\textbf{Theorem 2.} We optimize the reconstruction by minimizing the entropy $H(\mathbf{E} \mid \mathbf{E}_{ND}, \mathbf{E}_{ED})$. Ideally, when $H(\mathbf{E} \mid \mathbf{E}_{ND}, \mathbf{E}_{ED}) - \mathbb{E}_{\mathbf{E}, \mathbf{E}_{ND}, \mathbf{E}_{ED}}\left[\log p (\mathbf{E} \mid \mathbf{E}_{ND}, \mathbf{E}_{ED})\right]=0$, we achieve the best feature disentanglement. However, in practice, the estimation of conditional probability $p(\mathbf{E} \mid \mathbf{E}_{ND}, \mathbf{E}_{ED})$ is very tricky and complicated. Therefore, we use an approximate variational distribution $q(\mathbf{E} \mid \mathbf{E}_{ND}, \mathbf{E}_{ED})$ to simplify the calculation and optimization process. We provide a theoretical upper bound on $H(\mathbf{E} \mid \mathbf{E}_{ND}, \mathbf{E}_{ED})$ as follows:
\begin{gather}
	H(\mathbf{E} \mid \mathbf{E}_{ND}, \mathbf{E}_{ED})\leqslant \max\{\left\|\mathbf{E}-\mathbf{E}_{ND}\right\|_2^2, \left\|\mathbf{E}-\mathbf{E}_{ED}\right\|_2^2\}
\end{gather}

\textbf{\textit{Proof.}} For a given original view $\mathbf{E}$ and two augmented views $\mathbf{E}_{ND}$ and $\mathbf{E}_{ED}$, we have:
\begin{gather}
	\begin{aligned}p\left(\mathbf{E}_{ND},\mathbf{E}_{ED}\right)&=p\left(\mathbf{E}_{ND}\right)p\left(\mathbf{E}_{ED}\right)\\p\left(\mathbf{E}_{ND},\mathbf{E}_{ED}\mid \mathbf{E}\right)&=p\left(\mathbf{E}_{ND}\mid \mathbf{E}\right)p\left(\mathbf{E}_{ED}\mid \mathbf{E}\right)\end{aligned}
\end{gather}

\textbf{\textit{Lemma 1.}} For three given random variables $a, b, c$, if $p(b, c) = p(b)p(c)$ and $p(b, c\mid a) = p(b\mid a)p(c\mid a)$, then $I(a, b\mid c) = I(a, b)$. Based on the definition of mutual information, we deduce:
\begin{gather}
	\begin{aligned}
		&I\left(a;b\mid c\right)= \\
		&=\sum_{a}\sum_{b}\sum_{c}p\left(a,b,c\right)\log\frac{p\left(a,b,c\right)p\left(c\right)}{p\left(a,c\right)p\left(b,c\right)} \\
		&=\sum_{a}\sum_{b}\sum_{c}p\left(a\right)p\left(b,c\mid a\right)\log\frac{p\left(b,c\mid a\right)p\left(c\right)}{p\left(c\mid a\right)p\left(b\right)p\left(c\right)} \\
		&=\sum_{a}\sum_{b}\sum_{c}p\left(a\right)p\left(b\mid a\right)p\left(c\mid a\right)\log\frac{p\left(b\mid a\right)p\left(c\mid a\right)}{p\left(c\mid a\right)p\left(b\right)} \\
		&=\sum_{a}\sum_{b}p\left(a\right)p\left(b\mid a\right)\log\frac{p\left(b\mid a\right)}{p\left(b\right)} \\
		&=\sum_{a}\sum_{b}p\left(a,b\right)\log\frac{p\left(b\mid a\right)}{p\left(b\right)} \\
		&=I\left(a;b\right)
\end{aligned}
\end{gather}

According to Lemma 1, we can derive the theoretical bound of $I(\mathbf{E};\mathbf{E}_{ND},\mathbf{E}_{ED})$ as follows:
\begin{gather}
	\begin{aligned}
		&I\left(\mathbf{E};\mathbf{E}_{ND},\mathbf{E}_{ED}\right)
		\\& =I\left(\mathbf{E};\mathbf{E}_{ND}\mid\mathbf{E}_{ED}\right)+I\left(\mathbf{E};\mathbf{E}_{ND}\right)+I\left(\mathbf{E};\mathbf{E}_{ED}\right) \\
		&\geqslant I (\mathbf{E}_{ND}, \mathbf{E}_{ED};\mathbf{E}) \\
		&=I(\mathbf{E}_{ND};\mathbf{E})+I (\mathbf{E}_{ED};\mathbf{E}) \\
		&\geqslant I\left(\mathbf{E}_{ND};\mathbf{E}_{ED}\right)
\end{aligned}
\end{gather}

Assuming $q$ is a Gaussian distribution, the reconstruction process can be equivalent to minimizing the information entropy and its theoretical upper bound is formally defined as follows:
\begin{gather}
	H(\mathbf{E} \mid \mathbf{E}_{ND}, \mathbf{E}_{ED})\leqslant \max\{\left\|\mathbf{E}-\mathbf{E}_{ND}\right\|_2^2, \left\|\mathbf{E}-\mathbf{E}_{ED}\right\|_2^2\}
\end{gather}

\textbf{\textit{Proof.}} The estimation of conditional probability $p(\mathbf{E} \mid \mathbf{E}_{ND}, \mathbf{E}_{ED})$ is very tricky and complicated. Therefore, we use an approximate variational distribution $q(\mathbf{E} \mid \mathbf{E}_{ND}, \mathbf{E}_{ED})$ to simplify the calculation and optimization process. Therefore, we have,
\begin{gather}
	\begin{aligned}
	&H(\mathbf{E} \mid \mathbf{E}_{ND}, \mathbf{E}_{ED})\\
	&=-\mathbb{E}_{p(\mathbf{E} \mid \mathbf{E}_{ND}, \mathbf{E}_{ED})}\left[\log p\left(\mathbf{E} \mid \mathbf{E}_{ND}, \mathbf{E}_{ED}\right)\right] \\
	&\leq-\mathbb{E}_{p(\mathbf{E} \mid \mathbf{E}_{ND}, \mathbf{E}_{ED})}\left[\log q\left(\mathbf{E}\mid\mathbf{E}_{ND}, \mathbf{E}_{ED}\right)\right] \\
	&- D_{\mathrm{KL}}\left(p\left(\mathbf{E} \mid \mathbf{E}_{ND}, \mathbf{E}_{ED}\right)\|q\left(\mathbf{E} \mid \mathbf{E}_{ND}, \mathbf{E}_{ED}\right)\right)
	\end{aligned}
\end{gather}

Assume $q\left(\mathbf{E}\mid\mathbf{E}_{ND}, \mathbf{E}_{ED}\right)$ is a Gaussian distribution $\mathcal{N}\left(\mathbf{E}\mid \mathbf{E}_{ND}, \mathbf{E}_{ED},\sigma^2\mathbf{I}\right)$:
\begin{gather}
	\begin{aligned}
		&H(\mathbf{E} \mid \mathbf{E}_{ND}, \mathbf{E}_{ED})\\
		& \leqslant-\mathbb{E}_{p\left(\mathbf{E} \mid \mathbf{E}_{ND}, \mathbf{E}_{ED}\right)}\left[\log q\left(\mathbf{E} \mid \mathbf{E}_{ND}, \mathbf{E}_{ED}\right)\right] \\
		&=-\mathbb{E}_{p\left(\mathbf{E} \mid \mathbf{E}_{ND}, \mathbf{E}_{ED}\right)}\left[\log\left(\frac{1}{\sqrt{2\pi I}\sigma}e^{-\frac{1}{2}\frac{\left(\mathbf{E}-\mathbf{E}_{\{ND,ED\}}\right)^{2}}{(\sigma^{2}\mathbf{I})}}\right)\right] \\
		&=-\mathbb{E}_{p\left(\mathbf{E} \mid \mathbf{E}_{ND}, \mathbf{E}_{ED}\right)}\left[\log\left(\frac{1}{\sqrt{2\pi I}\sigma}\right)-\frac{\left(\mathbf{E}-\mathbf{E}_{\{ND,ED\}}\right)^{2}}{2\sigma^{2}\mathbf{I}}\right]
\end{aligned}
\end{gather}
Therefore, we get the upper bound of $H(\mathbf{E} \mid \mathbf{E}_{ND}, \mathbf{E}_{ED})$.

\section{Experiments}

\subsection{Experimental Setup}

\textbf{Datasets Description} \quad In our experiments, seven publicly available benchmark datasets are used including two Amazon items datasets \cite{shchur2018pitfalls} (i.e., Computers, and Photo), five citation network datasets \cite{yang2016revisiting} (i.e., Citeseer, Pubmed, DBLP, CoraFull, and Cora), three large-scale datasets \cite{hu2020open} (i.e., Ogbn-arxiv, Ogbn-mag, and Ogbn-products), page network datasets \cite{mernyei2020wiki, rozemberczki2021multi} (i.e., Wiki-CS and Croco).

\begin{table*}[htbp]
	\centering
	\belowrulesep=0pt
	\aboverulesep=0pt
	\caption{Experimental results on seven publicly available datasets. Classification accuracy (\%) is chosen as our evaluation metric. The best result in each column is in bold.}
	\renewcommand\arraystretch{1.2}
	\setlength{\tabcolsep}{2.3mm}{
		\begin{tabular}{l||ccccccc}
			\toprule[1pt]
			\rowcolor{gray!30}
			Methods     & Cora     & Citeseer  & PubMed   & Photo    & Computers & Ogbn-arxiv & Ogbn-products \\ \midrule \midrule
			Raw Feature \cite{mo2022simple} & 47.9$\pm$0.4 & 49.3$\pm$0.3  & 69.1$\pm$0.2 & 78.5$\pm$0.2 & 73.8$\pm$0.1  & 56.3$\pm$0.3   & 59.7$\pm$0.2      \\
			Deep Walk \cite{perozzi2014deepwalk}   & 81.5$\pm$0.2 & 43.2$\pm$0.4  & 65.3$\pm$0.5 & 89.4$\pm$0.1 & 85.3$\pm$0.1  & 63.6$\pm$0.4   & 73.2$\pm$0.2      \\
			GCN  \cite{kipfsemi}       & 81.5$\pm$0.2 & 70.3$\pm$0.4  & 79.0$\pm$0.5 & 91.8$\pm$0.1 & 84.5$\pm$0.1  & 70.4$\pm$0.3   & 81.6$\pm$0.4      \\
			GAT \cite{velivckovicgraph}         & 83.0$\pm$0.2 & 72.5$\pm$0.3  & 79.0$\pm$0.5 & 91.8$\pm$0.1 & 85.7$\pm$0.1  & \underline{70.6$\pm$0.3}   & {82.4$\pm$0.4}      \\
			GAE \cite{kipf2016variational}        & 74.9$\pm$0.4 & 65.6$\pm$0.5  & 74.2$\pm$0.3 & 91.0$\pm$0.1 & 85.1$\pm$0.4  & 63.6$\pm$0.5   & 72.1$\pm$0.1      \\
			VGAE  \cite{kipf2016variational}      & 76.2$\pm$0.4 & 66.7$\pm$0.5  & 75.7$\pm$0.3 & 91.4$\pm$0.1 & 85.7$\pm$0.3  & 64.8$\pm$0.2   & 72.9$\pm$0.2      \\
			DGI \cite{velivckovicdeep}        & 82.3$\pm$0.5 & 71.5$\pm$0.4 & 79.4$\pm$0.3 & 91.3$\pm$0.1 & \underline{87.8$\pm$0.2}  & 65.1$\pm$0.4   & 77.9$\pm$0.2      \\
			GMI \cite{peng2020graph}        & 83.0$\pm$0.2 & 71.5$\pm$0.4  & 79.9$\pm$0.4 & 90.6$\pm$0.2 & 82.2$\pm$0.4  & 68.2$\pm$0.2   & 76.8$\pm$0.3      \\
			GRACE  \cite{zhu2020deep}     & {83.1$\pm$0.2} & 72.1$\pm$0.1  & 79.6$\pm$0.5 & 91.9$\pm$0.3 & 86.8$\pm$0.2  & 68.7$\pm$0.4   & 77.4$\pm$0.4      \\
			MVGRL \cite{hassani2020contrastive}      & 82.9$\pm$0.3 & \underline{72.6$\pm$0.4}  & 80.1$\pm$0.7 & 91.7$\pm$0.1 & 86.9$\pm$0.1  & 68.1$\pm$0.1   & 78.1$\pm$0.1      \\
			GCA \cite{zhu2021graph}        & 81.8$\pm$0.2 & 71.9$\pm$0.4  & {81.0$\pm$0.3} & \underline{92.4$\pm$0.4} & 87.7$\pm$0.1  & 68.2$\pm$0.2   & 78.4$\pm$0.3      \\
			GIC \cite{mavromatis2021graph}        & 81.7$\pm$0.5 & 71.9$\pm$0.9  & 77.4$\pm$0.5 & 91.6$\pm$0.1 & 84.9$\pm$0.2  & 68.4$\pm$0.4   & 75.8$\pm$0.2      \\
			CRLC \cite{10036344}  &  \underline{83.5$\pm$ 0.2}  &  72.4$\pm$ 0.5   & \underline{82.0$\pm$ 0.1}  & 92.2$\pm$0.2 & 87.2$\pm$ 0.4   & 68.8$\pm$0.3  & \underline{82.6$\pm$ 0.3} \\
			NIGCN \cite{huang2023node}  &  83.4$\pm$ 0.3  &  71.6$\pm$ 0.3  & 81.6$\pm$ 0.3  & 91.6$\pm$ 0.2 &85.7$\pm$ 0.3    & 60.5$\pm$ 0.5  & 74.5$\pm$ 0.5 \\
			CGRL (Ours) &       \textbf{86.3$\pm$0.2} & \textbf{75.4$\pm$0.2}  & \textbf{84.5$\pm$0.6} & \textbf{93.7$\pm$0.3} & \textbf{89.7$\pm$0.5}  & \textbf{74.7$\pm$0.4}   & \textbf{85.1$\pm$0.3}   \\
			\bottomrule
	\end{tabular}}
\end{table*}

\begin{table*}[htbp]
	\centering
	\belowrulesep=0pt
	\aboverulesep=0pt
	\caption{Experimental results on five publicly available datasets. Classification accuracy (\%) is chosen as our evaluation metric. The best result in each column is in bold.}
	\renewcommand\arraystretch{1.2}
	\setlength{\tabcolsep}{4.5mm}{
	\begin{tabular}{l||ccccc}
		\toprule[1pt]
		\rowcolor{gray!30}
		Methods      & Wiki-CS  & DBLP     & Croco    & CoraFull & Ogbn-mag \\ \midrule \midrule
		Raw Feature \cite{mo2022simple}  & 72.0±0.9 & 71.6±0.6 & 41.7+0.4 & 43.6±0.7 & 22.1±0.3 \\
		DeepWalk \cite{perozzi2014deepwalk}    & 74.4±0.8 & 76.0±0.7 & 42.5±0.7 & 53.2±0.5 & 25.6±0.3 \\ 
		GCN  \cite{kipfsemi}        & 74.0±0.7 & 77.8±0.5 & 52.6±0.8 & 59.4+0.6 & 30.1±0.3 \\
		GAT  \cite{velivckovicgraph}        & 77.6±0.6 & 78.2±1.5 & 53.3+1.0 & 58.6±0.5 & 30.5±0.3 \\
		DGI  \cite{velivckovicdeep}        & 74.8±0.7 & 83.1±0.5 & 53.1±0.7 & 55.1±0.6 & 30.6±0.3 \\
		GIC  \cite{mavromatis2021graph}         & 75.9±0.6 & 81.9±0.8 & 56.8+0.6 & 58.2±0.7 & 29.8±0.2 \\
		GRACE \cite{zhu2020deep}       & 75.3±0.7 & 84.2±0.6 & 58.3±0.4 & 54.0±0.6 & 31.1±0.3 \\
		GMI \cite{peng2020graph}          & 74.8±0.7 & 83.9±0.8 & 54.3±0.9 & 54.6+0.8 & 27.2±0.1 \\
		MVRLG  \cite{hassani2020contrastive}      & 76.3±1.1 & 79.5±0.8 & 57.9±0.6 & 58.8±0.7 & 30.4±0.4  \\
		Contrast-Reg \cite{ma2021improving} & 77.0±0.6 & 83.6±0.8 & 58.4±0.7 & 58.9±0.6 & 30.9±0.4 \\
		GRLC         & \underline{77.9±0.5} & \underline{84.2±0.6} & \underline{59.5±0.7} & \underline{59.4±0.6} & \underline{31.6±0.2} \\ 
		CGRL (Ours)  & \textbf{80.4±0.3} & \textbf{87.2±0.7} & \textbf{63.6±0.4} & \textbf{62.9±0.5} & \textbf{35.6±0.4}
		\\ \bottomrule
	\end{tabular}}
\end{table*}

\begin{figure*}[htbp]
	\centering
	\includegraphics[width=0.98\linewidth]{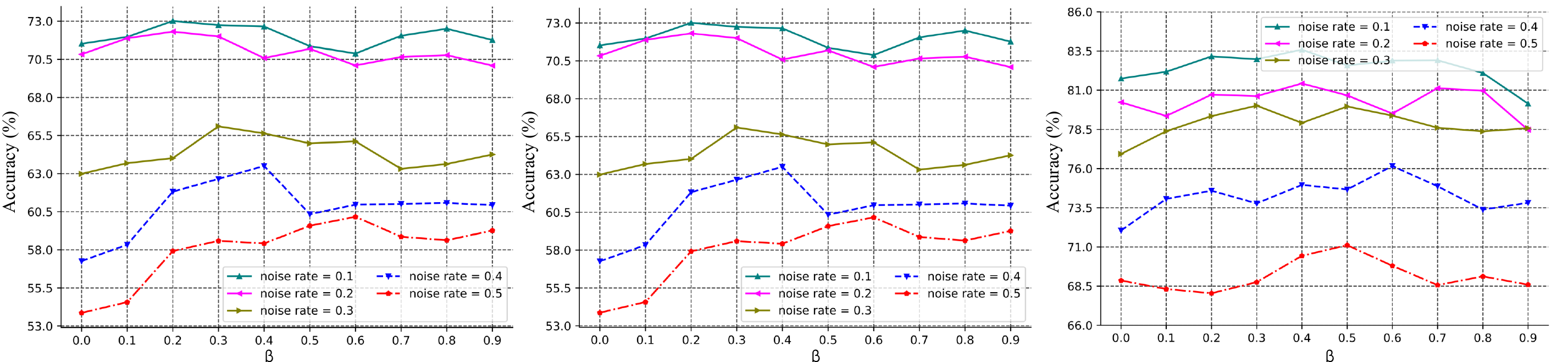}
	\caption{The impact of different noise rates on experimental results on Cora, Citeseer and PubMed datasets.}
	\label{fig:noise}
\end{figure*}

\textbf{Evaluation Metrics} \quad We use classification accuracy to evaluate the performance of our method CGRL and other comparative methods.

\textbf{Comparison Methods} \quad
We compare our method CGRL with twelve state-of-the-art deep learning-based algorithms, including two traditional graph embedding algorithms (i.e., raw features \cite{mo2022simple}, and DeepWalk \cite{perozzi2014deepwalk}), three semi-supervised algorithms (i.e., GCN \cite{kipfsemi}, NIGCN\cite{huang2023node}, and GAT \cite{velivckovicgraph}), and nine self-supervised algorithms (i.e., GAE \cite{kipf2016variational}, VGAE \cite{kipf2016variational}, DGI \cite{velivckovicdeep}, GCA \cite{zhu2021graph}, MVGRL \cite{hassani2020contrastive}, GIC \cite{mavromatis2021graph}, GRACE \cite{zhu2020deep}, GMI \cite{peng2020graph}, and CRLC \cite{10036344}).

\textbf{Setting-up} 
All the experiments in this paper are implemented on a server with 2 A100 (total 160GB memory). For each experiment, we run the code five times with a random seed to obtain the final mean and corresponding standard deviation to avoid experimental chance. Futhermore, for some parameter settings of the model, we set epochs to 1000/300, batch-size to full-batch/mini-batch, learning rate to 0.005, mask rate to 0.5, $\alpha$ to 0.5, $\beta$ is 0.5, the activation function to GELU, dropout is 0.2, the weight decay to 1e-4, learning rate scheduling to cosine, warmup epochs to 100, and hidden\_size to 128. We utilize the Adam optimization algorithm to update parameters.

\subsection{Results and Analysis}

\begin{table*}[htbp]
	\centering
	\belowrulesep=0pt
	\aboverulesep=0pt
	\caption{Ablation studies are performed on seven datasets to verify the effectiveness of node-masking view, edge perturbation view, and information bottleneck strategy. Classification accuracy (\%) is chosen as our evaluation metric. The best result in each column is in bold.}
	\label{tab:2}
	\renewcommand\arraystretch{1.2}
	\setlength{\tabcolsep}{2.5mm}{
		\begin{tabular}{ccc||ccccccc}
			\toprule
			\rowcolor{gray!30}
			$\mathcal{L}_{CLS}^{ND}$      & $\mathcal{L}_{CLS}^{ED}$                             & $I(E, \tilde{E})$                             & Cora & Citeseer & PubMed & Photo & Computers & Ogbn-arxiv & Ogbn-products \\ \midrule \midrule
			\XSolidBrush   & \XSolidBrush   & \XSolidBrush   &   83.0   &   72.5       &    79.0    &   91.8    &   84.5        &    70.4        &   81.6            \\
			\CheckmarkBold & \XSolidBrush   & \XSolidBrush   &   83.7   &   72.7       &   79.5     &  92.1     &  85.3         &   70.8         &    82.0           \\
			\XSolidBrush   & \CheckmarkBold & \XSolidBrush   & 83.2     &  72.5        &  79.3      &  92.0     &    84.7       &   70.5         &   81.6            \\
			\CheckmarkBold & \CheckmarkBold & \XSolidBrush   & 84.0     &  72.9        &  81.8      &  92.4     &   86.6        &    71.5        &     82.4          \\
			\CheckmarkBold & \XSolidBrush   & \CheckmarkBold &  84.0    &    73.1      &     82.3   &    92.6   &   88.1        &   71.9         &  83.0             \\
			\XSolidBrush   & \CheckmarkBold & \CheckmarkBold &  83.8    &     72.8     &    80.0    & 92.2      &   85.6        &   70.8         &  82.1             \\
			\CheckmarkBold & \CheckmarkBold & \CheckmarkBold &   \textbf{86.3} & \textbf{75.4}  & \textbf{84.5} & \textbf{93.7} & \textbf{89.7}  & \textbf{74.7}   & \textbf{85.1}             \\ \bottomrule
	\end{tabular}}
\end{table*}

\begin{table*}[htbp]
	\centering
	\belowrulesep=0pt
	\aboverulesep=0pt
	\caption{Experiments were conducted using different graph convolutional neural networks on seven data sets to verify their impact on experimental results. Classification accuracy (\%) is chosen as our evaluation metric. The best result in each column is in bold.}
	\label{tab:3}
	\renewcommand\arraystretch{1.2}
	\setlength{\tabcolsep}{3.2mm}{
		\begin{tabular}{c||ccccccc}
			\toprule
			\rowcolor{gray!30}
			GraphConv & Cora & Citeseer & PubMed & Photo & Computers & Ogbn-arxiv & Ogbn-products \\ \midrule \midrule
			GCN \cite{kipfsemi}                             &  82.7    &   70.7       &  81.4      &  88.6     & 86.1          &   70.9         & 82.3              \\
			GAE \cite{kipf2016variational}                              &   79.3   &  68.4        &    76.8    &  90.6     & 85.9          &  67.2          &   74.6            \\
			VGAE \cite{kipf2016variational}                  &     79.8    &   71.2   &   77.4         & 91.2             &   86.6        &    68.1        & 77.4              \\
			GIN  \cite{xu2018powerful}                            &   83.3   &   71.8       &  81.2      &  90.5     &   86.9        &   71.3         &    82.6           \\
			GraphSAGE  \cite{hamilton2017inductive}                      &    83.6  &     72.8     &   82.0     &   92.4    &    87.6       &   71.9         &    82.8           \\
			GAT \cite{velivckovicgraph}                &             \textbf{86.3} & \textbf{75.4}  & \textbf{84.5} & \textbf{93.7} & \textbf{89.7}  & \textbf{74.7}   & \textbf{85.1}               \\  \bottomrule
	\end{tabular}}
\end{table*}

\textbf{Node classification.}
Tables 1 and 2 summarize the node classification accuracy of the baselines and the proposed method CGRL on twelve real graph-structured data sets. Specifically, our approach leverages traditional graph embedding algorithms (i.e., raw features and DeepWalk). For example, our method CGRL improves the average accuracy by 20.31\% and 10.76\% compared to raw features and DeepWalk methods respectively. Compared with self-supervised methods (e, g., DGI, and GCA, etc.), CGRL also achieves better performance. In addition, CGRL also outperforms semi-supervised algorithms (i.e., GCN , NIGCN, and GAT). The performance improvement may be attributed to the design of the multi-view contrast learning strategy for adaptive masking nodes and perturbed edges, which enables the model to automatically learn whether to mask nodes and perturbed edges. To ensure that our model can reconstruct nodes and edges, we use a random walk strategy to sample the node's subgraph structure to blur its representation. However, most other baselines perform multi-view contrastive learning by randomly dropping nodes and edges, which will seriously destroy the semantic information of the graph. In addition, we also introduce the information bottleneck theory to ensure that the augmented views are structurally heterogeneous but semantically similar. The intuition behind this is that maximizing the mutual information between multiple views will lead to a consistent representation of the augmented views learned by the model, which leads to overfitting of the model. 

\begin{figure}
	\centering
	\includegraphics[width=1\linewidth]{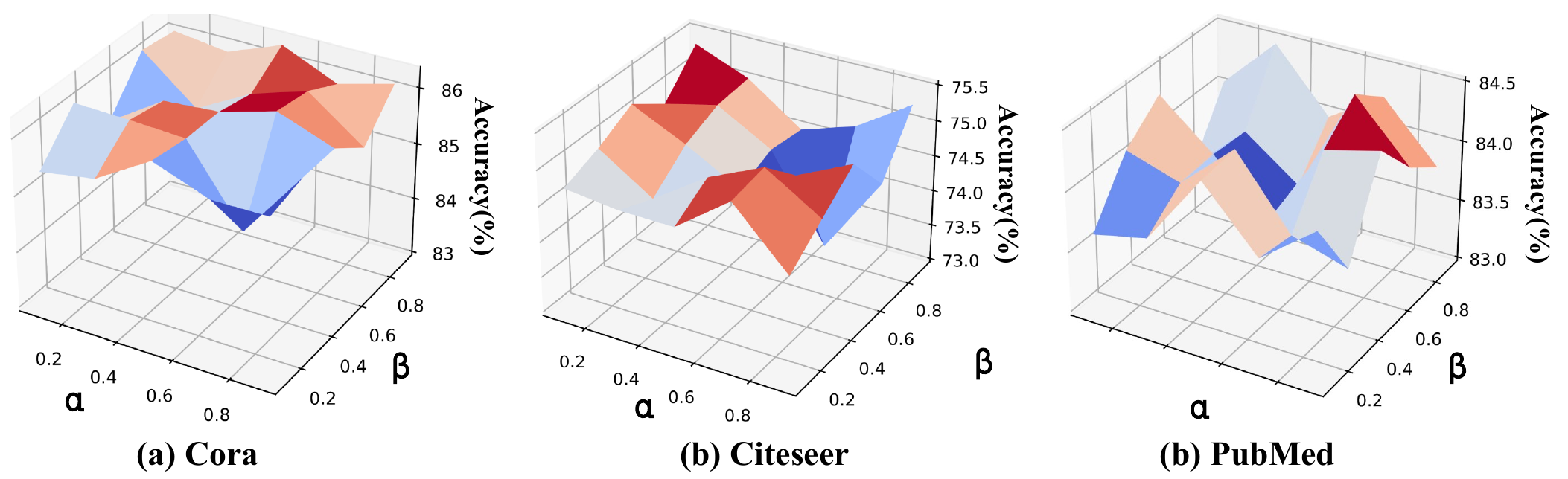}
	\caption{The impact of hyperparameter settings (i.e., $\alpha$ and $\beta$) on experimental results in four data sets (i.e., Cora, Citeer, and PubMed).}
	\label{fig:hyparameter}
\end{figure}

\textbf{Effect of noise.} In order to verify the anti-interference ability of our model CGRL, we perform experiments to demonstrate the performance by varying the noise rate in $\{0, 0.1, 0.2, 0.3, 0.4, 0.5\}$ and the $\beta$ in $\{0.1, 0.2, 0.3, 0.4, 0.5\\,0.6, 0.7, 0.8, 0.9\}$. The experimental results are shown in Fig. \ref{fig:noise}, We can find that when the parameter $\beta$ is 0, CGRL has the worst effect on the Cora, Citeseer and PubMed data sets. When $\beta$ is greater than 0, CGRL's anti-interference ability is significantly reflected, and the best effect is in the range of 0.2 to 0.6. The experimental results show that the introduction of information bottleneck theory can enhance the noise invariance ability of the model.

\textbf{Hyper-parameter analysis.}
We set two hyperparameters $\alpha$ and $\beta$ during the model optimization process. Their settings will have a relatively large impact on the performance of the model. Therefore, we also investigated the impact of hyperparameters on CGRL. We performed the node classification task by varying $\alpha$ and $\beta$ from 0.1 to 0.9 and visualized the accuracy in Fig. \ref{fig:hyparameter}. When $\alpha$ and $\beta$ are set to relatively large values, the node classification effect of the model is better, and when set to smaller values, the performance is poor. The difference in performance can be attributed to the inability of smaller parameter settings to eliminate redundant information in the graph structure and to obtain optimal augmented multi-view.

\subsection{Ablation Study}
We performed three ablation experiments to verify the effectiveness of our proposed node-masking view, edge perturbation view, and information bottleneck theory.

\begin{figure*}[htbp]
	\centering
	\includegraphics[width=1\linewidth]{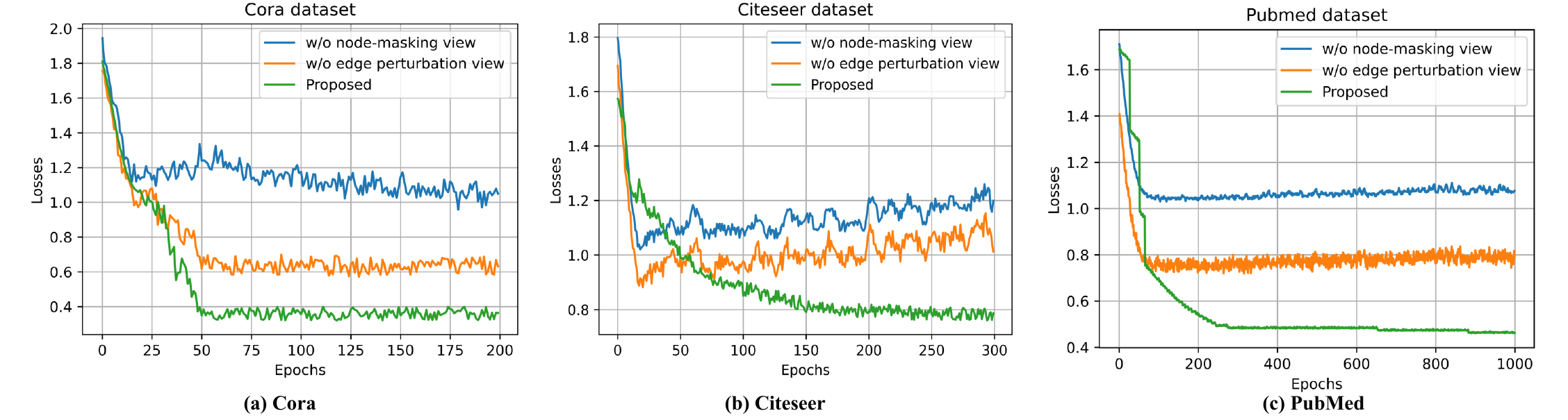}
	\caption{Effect of node-masking view and edge perturbation view on training loss.}
	\label{fig:loss}
\end{figure*}

\subsubsection{Effectiveness of adaptive graph contrastive learning via IB.}

We perform ablation studies to analyze the impact of node-masking view, edge perturbation view, and IB criterion on experimental results. The experimental results are shown in Table \ref{tab:2}. Firstly, when CGRL includes all modules, the accuracy of node classification on seven data sets is the highest. Secondly, We find that the node-masking view has a greater impact on the experimental results than the edge perturbation view. Experimental phenomena show that node classification mainly depends on the features of nodes. Thirdly, node-mask view and edge perturbation view combined with IB accuracy can further improve the accuracy of node classification. The performance improvement may be attributed that the IB criterion can promote multi-views to obtain structurally heterogeneous but semantically similar graph structures.

\subsubsection{Impact of GNN variants.} 
We explore the impact of different graph neural network variants on experimental results. As shown in Table \ref{tab:3}, GAT has the best effect on seven graph structure data sets, with node classification accuracy rates of 86.3\%, 75.4\% and 84.5\% on the Cora, Citeseer, and PubMed data sets, respectively, followed by GraphSAGE, with node classification accuracy rates of 83.6\%, 72.8\% and 82.0\% , respectively, and GCN has the worst effect, with node classification accuracy rates of 82.7\%, 70.7\% and 81.4\%, respectively. Therefore, we choose GAT as our encoder in our experiments.

\subsubsection{Effectiveness of multi-view optimization.}
We perform ablation experiments to verify the impact of multi-view augmentation on the optimization process of model training. The experimental results are shown in Fig. \ref{fig:loss}. On Cora, Citeseer, and PubMed datasets, we found that in the absence of node-masking view or edge perturbation view, the loss value of the model cannot converge to the optimal value, and the edge perturbation view has the worst convergence effect. When CGRL combines the node-masking view, edge perturbation view and information bottleneck criterion, the loss value of the model can converge to close to 0. The experimental results show that the performance of the augmented node-masking view is better than the augmented edge perturbation view, but worse than the augmented multi-view that combines information bottlenecks, which shows that the information bottleneck theory can further enhance the utilization of information.

\subsubsection{Visualization.}
\begin{figure}
	\centering
	\includegraphics[width=1\linewidth]{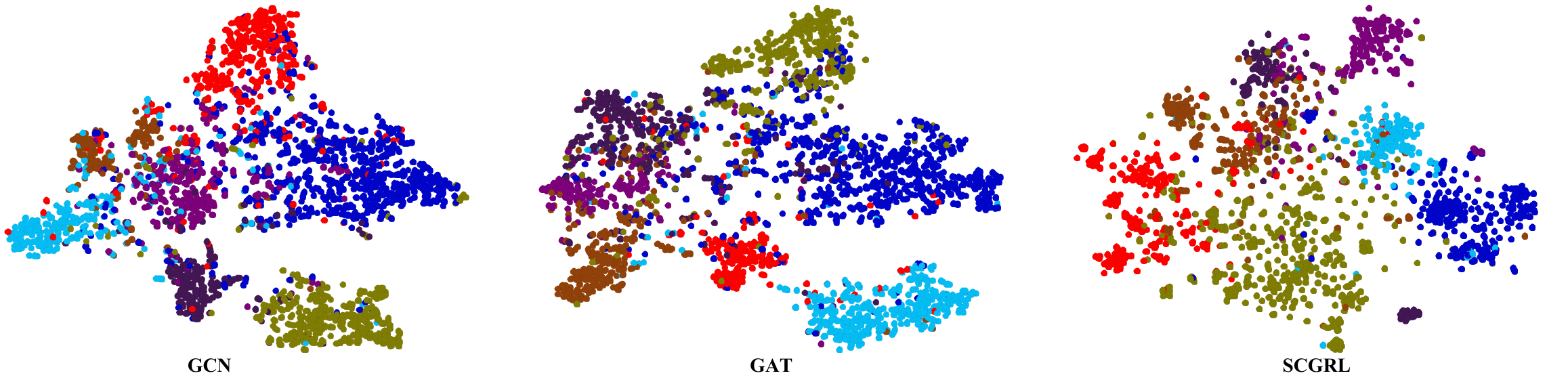}
	\caption{Visualization of feature embeddings in 2-dimensional space for different comparison algorithms on the Cora dataset.}
	\label{fig:cora-visual}
\end{figure}

\begin{figure}
	\centering
	\includegraphics[width=1\linewidth]{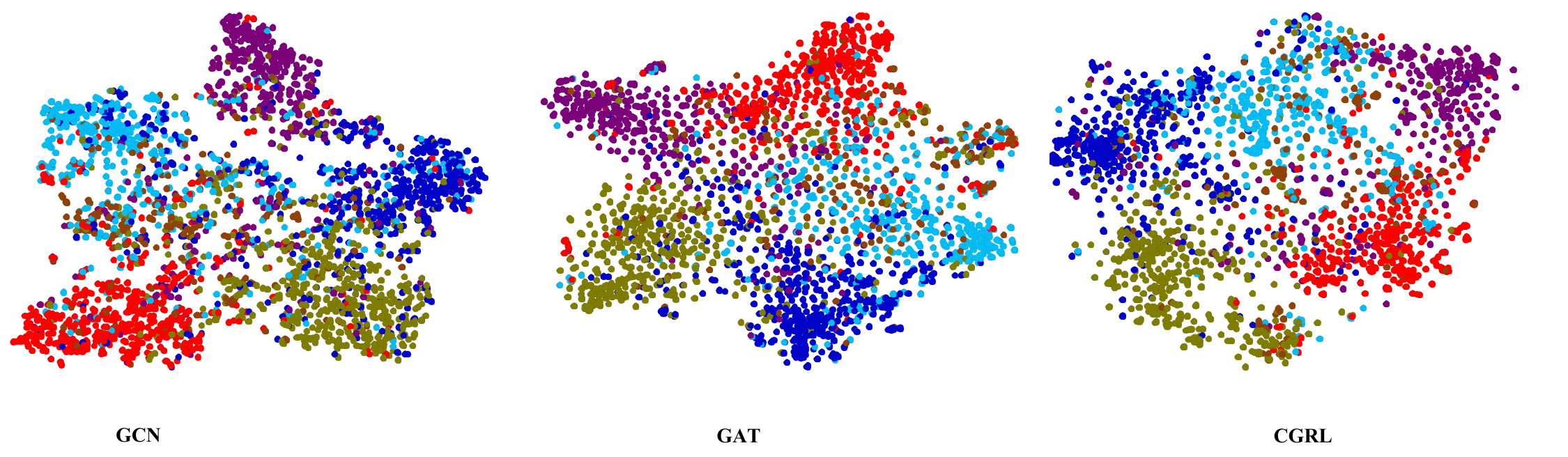}
	\caption{Visualization of feature embeddings in 2-dimensional space for different comparison algorithms on the Citeer dataset.}
	\label{fig:citeer-visual}
\end{figure}

We use the T-SNE method to project the high-dimensional node features into a 2-dimensional space and visualize them. The visualization results are shown in Figs. \ref{fig:cora-visual} and \ref{fig:citeer-visual}. In the Cora data set, GCN and GAT have more overlaps between different node categories and the class boundaries are not clear enough, while CGRL has clearer class boundaries between different node categories. In the Citeseer data set, the feature embeddings of GCN and GAT are more scattered among the same node categories, and there is excessive overlap between different categories, while CGRL is more densely distributed for the same category of nodes and there is less overlap between different categories.

\section{Conclusions}
In this paper, we propose a Contrastive Graph Representation Learning with Adversarial Cross-view Reconstruction and Information Bottleneck for node classification to automatically generate structurally heterogeneous but semantically similar multi-views. Specifically, CGRL can adaptively learn to mask nodes and perturb edges in the graph to obtain optimal graph structure representation. Furthermore, we innovatively introduce the information bottleneck theory into GCL to eliminate redundant information in multiple contrasting views while retaining as much information about node classification as possible. Moreover, we add noise perturbations to the original views and reconstruct the augmented views by constructing adversarial views to improve the robustness of node feature representation. Extensive experiments on real-world public datasets show that our approach significantly outperforms existing the SOTA algorithms.


\bibliographystyle{elsarticle-num}
\bibliography{refs}


%\begin{thebibliography}{00}
%%% Text of bibliographic item
%
%\bibitem{}
%
%\end{thebibliography}

\end{document}